\documentclass[runningheads]{llncs}
\usepackage{graphicx,dblfloatfix}
\newcommand{\orcidmanuel}	{\href{https://orcid.org/0000-0002-8510-1972}{\protect\includegraphics[scale=0.05]{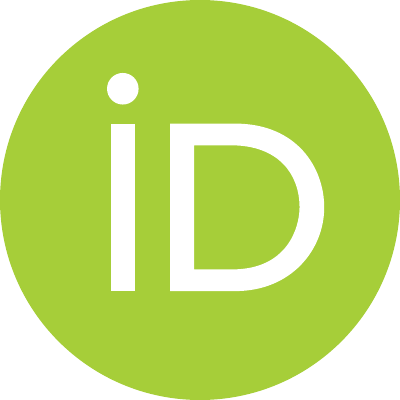}}}
\newcommand{\orcidoscar}	{\href{https://orcid.org/0000-0002-8296-6620}{\protect\includegraphics[scale=0.05]{images/orcid}}}
\newcommand{\orcidmarlon}	{\href{https://orcid.org/0000-0002-9247-7476}{\protect\includegraphics[scale=0.05]{images/orcid}}}

\usepackage{makeidx}  
\usepackage{booktabs} 
\usepackage{amsmath,amssymb,amsfonts}
\usepackage{nccmath}
\usepackage{float}
\usepackage{comment} 
\usepackage[table,xcdraw]{xcolor}
\usepackage{csquotes}
\usepackage[ruled]{algorithm2e} 
\usepackage{wrapfig}
\usepackage{subcaption}
\usepackage{hyperref}
\usepackage{multirow}
\usepackage[normalem]{ulem}
\useunder{\uline}{\ul}{}
\usepackage{cite}
\usepackage{algorithmic}
\usepackage{graphicx}
\usepackage{textcomp}
\usepackage{footnote}
\makesavenoteenv{tabular}
\makesavenoteenv{table}

\makeatletter
\newcommand\footnoteref[1]{\protected@xdef\@thefnmark{\ref{#1}}\@footnotemark}
\makeatother

\begin{document}
\title{Learning Accurate Business Process Simulation Models from Event Logs via Automated Process Discovery and Deep Learning\thanks{Work funded by European Research Council (PIX Project).}}
\titlerunning{Learning Accurate Hybrid Simulation Models of Business Processes }
%
\author{Manuel Camargo\inst{1,2,3}~\orcidmanuel \and Marlon Dumas\inst{1}~\orcidmarlon \and Oscar Gonz{\'a}lez-Rojas\inst{2}~\orcidoscar}
\authorrunning{M. Camargo et al.}
%
\institute{University of Tartu, Tartu, Estonia, \email{marlon.dumas@ut.ee}
\and
Universidad de los Andes, Bogot{\'a}, Colombia, \email{o-gonza1@uniandes.edu.co}
\and 
Apromore, Tartu, Estonia, 
\email{manuel.camargo@apromore.com}
}
\maketitle              
\setcounter{footnote}{0}

\begin{abstract}
Business process simulation is a well-known approach to estimate the impact of changes to a process with respect to time and cost measures -- a practice known as what-if process analysis. The usefulness of such estimations hinges on the accuracy of the underlying simulation model. Data-Driven Simulation (DDS) methods leverage process mining techniques to learn process simulation models from event logs. Empirical studies have shown that, while DDS models adequately capture the observed sequences of activities and their frequencies, they fail to accurately capture the temporal dynamics of real-life processes. In contrast, generative Deep Learning (DL) models are better able to capture such temporal dynamics. The drawback of DL models is that users cannot alter them for what-if analysis due to their black-box nature. This paper presents a hybrid approach to learn process simulation models from event logs wherein a (stochastic) process model is extracted via DDS techniques, and then combined with a DL model to generate timestamped event sequences. An experimental evaluation shows that the resulting hybrid simulation models match the temporal accuracy of pure DL models, while partially retaining the what-if analysis capability of DDS approaches.

\end{abstract}
\section{Introduction}
\label{sec:intro}

Business Process Simulation (BPS) models allow analysts to estimate the impact of changes to a process with respect to temporal and cost measures -- a practice known as ``what-if'' process analysis~\cite{FBPM}. However, the construction and tuning of BPS models is error-prone, as it requires careful attention to numerous  pitfalls~\cite{VanderAalst2015}. Moreover, the accuracy of manually tuned BPS models is limited by the completeness of the process model used as a starting point, yet manually designed models often do not capture exceptional paths. 
Previous studies have proposed to extract BPS models from execution data (event logs) via process mining techniques~\cite{Martin2016}. While Data-Driven Simulation (DDS) models extracted in this way can be tuned to accurately capture the control-flow and temporal behavior of a process~\cite{Camargo2020}, they suffer from fundamental limitations stemming from the expressiveness of the modeling notation (e.g.\ BPMN, Petri nets) and from assumptions about the resources' behavior. One such assumption is that all waiting times are due to resource contention (i.e.\ a resource not starting a task because it is busy with another task). Another assumption is that resources exhibit robotic behavior: if a resource is available, and it may perform an enabled activity instance, the resource will immediately start it. In other words, these approaches do not take into account behaviors such as multitasking, batching, fatigue effects, and inter-process resource sharing, among others~\cite{Estrada-Torres2021}.

Other studies have shown that Deep Learning (DL) generative models trained from logs can accurately predict the next event in a case and its timestamp or the suffix of a case starting from a given prefix~\cite{Tax2017,Evermann2017}. Suitably trained DL generative models can also be used to generate entire traces and even entire logs~\cite{Camargo2019}, which effectively allows us to use a DL generative model as a simulation model. Camargo~et~al.~\cite{Camargo2020ddsdl} empirically show that DL models are more accurate than DDS models when it comes to generating logs consisting of activity sequences with start and end timestamps. In particular, generative DL models can emulate delays between activities that DDS models do not capture. However, unlike DDS models, DL models are not suitable for what-if analysis due to their black-box nature -- they do not allow to specify a change to the process and to simulate the effect of this change.

This paper presents a method, namely DeepSimulator, that combines DDS and DL methods to discover a BPS model from a log. The idea is to use an automated process discovery technique to extract a process model with branching probabilities (a.k.a.\ stochastic process model~\cite{Leemans2021}) and to delegate the generation of activity start and end times to a DL model. 

The paper is structured as follows. Section~\ref{sec:background} discusses methods to learn generative models from logs using DDS and DL techniques. Section~\ref{sec:tool} presents the proposed method while Section~\ref{sec:evaluation} presents an empirical evaluation thereof. Finally, Section~\ref{sec:conclusion} draws conclusions and sketches future work.

%
\section{Related Work}
\label{sec:background}

\subsection{Data-driven simulation of business processes}

Previous studies on DDS methods can be classified in two categories. 
A subset of previous studies have proposed conceptual frameworks and guidelines to manually derive, validate, and tune BPS parameters from event logs~\cite{Wynn2008,Martin2016}, without seeking to automate the extraction process.  
Other studies have proposed methods that  automate the extraction and/or tuning of simulation parameters from logs. In this paper, we focus on automated methods. One of the earliest such methods is that of Rozinat~et~al.~\cite{Rozinat2009}, who propose a semi-automated approach to extract BPS models based on Colored Petri Nets. Later, Khodyrev~et~al.~\cite{Khodyrev2014} proposed an approach to extract BPS models from data, although they leave aside the resource perspective (i.e.\ the discovery of resource pools). 
More recently, Pourbafrani~et~al.~\cite{Pourbafrani2020} present an approach for generating DDS models based on time-aware process trees. In all of the above studies, the responsibility for tuning these parameters is left to the user. This limitation is addressed in the Simod method~\cite{Camargo2020}, which automates the extraction of BPS models by employing Bayesian optimization to tune the hyperparameters used to discover the process model and resource pools as well as the statistical parameters of the BPS model (branching probabilities, activity processing times, and inter-case arrival times). This tuning phase seeks to optimize the similarity between the logs produced by the extracted BPS model and (a testing fold of) the original log.

\subsection{Generative DL models of business processes}

A Deep Learning (DL) model is a network of  interconnected layers of neurons (perceptrons) that collectively perform non-linear data transformations~\cite{Lecun2015}. The objective of these transformations is to train the network to learn the patterns observed in the data. In theory, the more layers of neurons in the system, the more it will detect higher-level patterns via composition of complex functions~\cite{Lecun2015}. A wide range of neural network architectures have been proposed, e.g.\ feed-forward networks, Convolutional Neural Networks, Variational Auto-Encoders, Generative Adversarial Networks (GAN) (often in a combination with other architectures), and Recurrent Neural Networks (RNN). The latter type of architecture is specifically designed to handle sequential data.

DL models have been widely applied in the field of predictive process monitoring. Evermann et al.~\cite{Evermann2017} proposed an RNN architecture to generate the most likely remaining sequence of events (suffix) of an ongoing case. This architecture cannot handle numeric features and thus cannot generate timestamped events (timestamps are numeric). This limitation is shared by the approach in~\cite{lin2019mm} and others reviewed in~\cite{Tax2018}. Tax et al.~\cite{Tax2017} use an RNN architecture known as Long-Short-Term Memory (LSTM) to predict the next event in an ongoing case and its timestamp, and to generate the remaining sequence of timestamped events from a given prefix of a case. However, this approach cannot handle high-dimensional inputs due to its reliance on one-hot encoding of categorical features. Its precision deteriorates as the number of categorical features increases. This limitation is lifted by DeepGenerator~\cite{Camargo2019}, which extends the approach in~\cite{Tax2017} with two mechanisms to handle high-dimensional input: n-grams and embeddings. This approach also addresses the problem of generating long suffixes (not well handled in~\cite{Tax2017}) and entire traces, by using a random next-event selection approach. It is also able to associate a resource to each event in a trace. More recently, Taymouri et al.~\cite{Taymouri2020} proposed a GAN-LSTM architecture to train generative models that produce timestamped activity sequences (without associated resources).

Camargo et al.~\cite{Camargo2020ddsdl} compare the relative accuracy of DL models against DDS models for generating sequences of the form (activity, start timestamp, end timestamp). 
This comparison suggests that DL models may outperform DDS models when trained with large logs, while the opposite holds for smaller logs. 
Camargo et al.~\cite{Camargo2020ddsdl} additionally show that DL models generally outperform DDS methods when it comes to predicting activity start and end timestamps.

\section{Hybrid Learning of BPS Models}
\label{sec:tool}

\begin{figure*}[ht]
  \begin{center}
    \includegraphics[width=\textwidth]{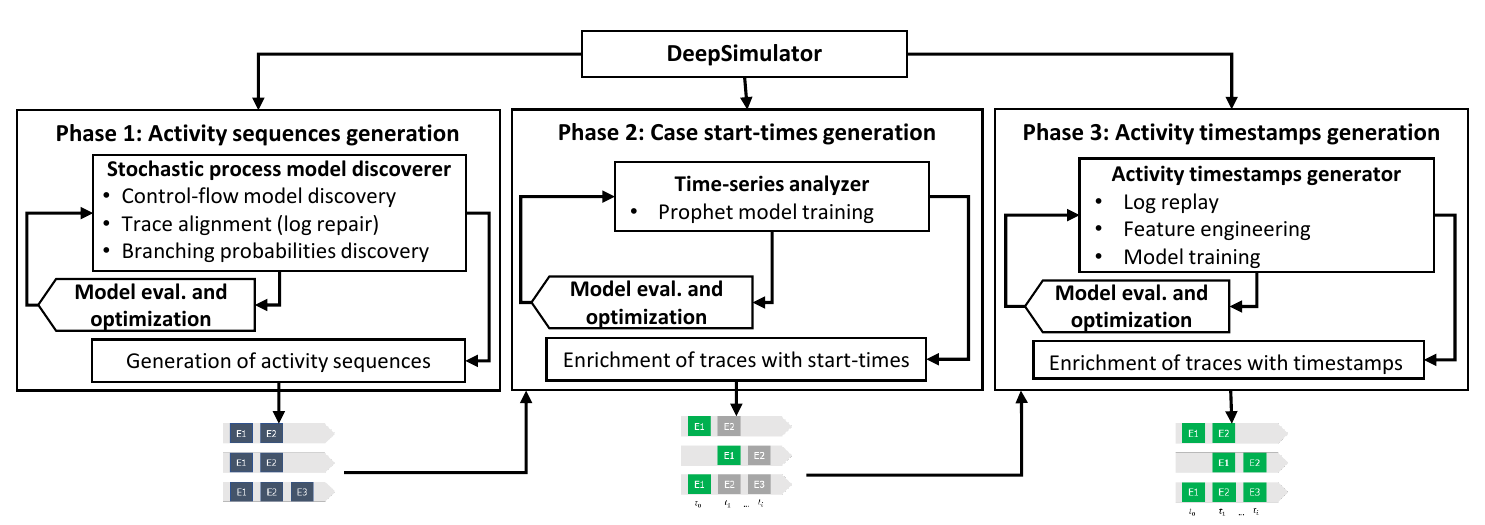}
    \caption{Overview of the proposed BPS model discovery method}
    \label{fig:architecture}
   \end{center}
    \vspace*{-10mm}
\end{figure*}

Fig.~\ref{fig:architecture} depicts the architecture of the DeepSimulator approach. The architecture is a pipeline with three phases. The first phase uses PM techniques to learn a model to generate sequences of (non-timestampted) events. The second and third phases enrich these sequences with case start times and activity start and end times. Below, we discuss each phase in turn.

\paragraph{Phase 1: Activity sequences generation.} \label{sec:hb_tool_phase1}
The aim of this phase is to extract a stochastic process model~\cite{Leemans2021} from the log and to use it to generate sequences of activities that resemble those in the log. A stochastic process model is a process model with branching probabilities assigned to each branch of a decision point. In this paper, we represent process models using the standard BPMN notation. Phase~1 starts with a \emph{control-flow discovery} step, where we first discover a plain (non-stochastic) process model using the Split Miner algorithm~\cite{Augusto2017}.
This algorithm relies on two parameters: the sensitivity of the parallelism oracle ($\eta$) and the level of filtering of directly-follow relations ($\epsilon $). The former parameter determines how likely the algorithm will discover parallel structures, while the latter determines the percentage of directly-follows relations between activity types are captured in the resulting model.
Like other automated process discovery algorithms, the Split Miner discovers a process model that does not perfectly fit the log.
The discovered process model cannot parse some traces in the log. This hinders the calculation of the branching probabilities.
Accordingly, we apply the \emph{trace alignment} algorithm in~\cite{DBLP} to compute an alignment for each trace in the log that the model cannot parse. An alignment describes how a trace can be modified to be turned into a trace that can be parsed by the model (via ``skip'' operations). 
Based on the alignments, we either \emph{repair} each non-conformant trace, or we \emph{replace} it with a copy of the most similar conformant trace (w.r.t.\ string-edit distance). The choice between the repair and the replacement approaches is a parameter of the method. 

Next, DeepSimulator uses the (conformant) event log to discover the \emph{branching probabilities} for each branching point in the model. Here, DeepSimulator offers two options: (i) assign equal values to each conditional branch; or (ii) compute the branching probabilities by replaying the aligned event against the process model. The first approach may perform better for smaller logs, where the probabilities computed via replay are not always reliable, while the latter may be preferable for larger logs. 

The DeepSimulator combines the process model and the branching probabilities to assemble a stochastic process model. In this step, the DeepSimulator uses a Bayesian optimization technique to discover the hyperparameter settings (i.e., values of $\epsilon$, $\eta$, replace-vs-repair, and equal-vs-computed probabilities) that maximize the similarity between the generated and the ground truth sequences in terms of activity sequences. The optimizer uses a holdout method, and as a loss function, it uses the Control-Flow Log Similarity (CFLS) metric described in~\cite{Camargo2020}. The CFLS metric is the mean string-edit distance between the activity sequences generated by the stochastic process model and the traces in the ground-truth log after their optimal alignment.\footnote{We did not use the stochastic conformance checking metrics over Petri nets of~\cite{Leemans2021} since our method handles BPMN models with inclusive join gateways, which cannot be directly transformed to Petri Nets (without exponential blowout) as shown in~\cite{Favre2012}}
Finally, in the \emph{sequences' generation} step, DeepSimulator uses the resulting stochastic process model to generate a bag of activity sequences without timestamps. This bag is used as the log's base structure in Phase~3.

\paragraph{Phase 2: Case start-times generation.} \label{sec:hb_tool_phase2}
In this phase, we generate each process instance's start time in the output log. Traditionally, DDS models generate the start-time of cases by randomly drawing from an unimodal distribution of the interarrival times between consecutive cases. A typical BPS model captures the interarrival times using a negative exponential distribution (i.e., it models the creation of cases as a Poisson process). However, a single distribution is not realistic enough to capture real scenarios. For example, cases might be created more frequently on Mondays than on Thursdays in a claims handling process.

Instead of fitting an interarrival distribution, the DeepSimulator models the case generation as a time series prediction problem as the number of cases generated per hour of the day. This type of modeling allows us to use robust techniques such as ARIMA or ETS tested successfully in several contexts such as Stock Market Analysis or Workload Projections. DeepSimulator uses the Prophet~\cite{Taylor2018} model proposed by Facebook because it is one of the simplest but, at the same time, more accurate predictive models for this type of task. Prophet starts from the time series decomposition into four main components (i.e., trend, seasonality, holidays, and error) and applies specialized techniques to model each component. 

The trend component decomposes those non-periodic changes in the time series values, which are modeled using logistic growth models or Piecewise linear models. The seasonality component decomposes the periodic changes repeated at fixed intervals (hours, weeks, months, or years), which are modeled by using the Fourier series. The holidays component represents the effects of holidays that occur on potentially irregular schedules over one or more days. This component is optionally modeled and is defined manually by a domain expert, since it is specific to each time series. The model automatically calculates the error, corresponding to all those unforeseen changes that the model cannot fit.

In the \emph{time-series analysis} step, we use a saturated logistic growth model to fit the case generation trend. We chose this model, considering that the time series is limited by a lower and upper bound. The lower bound corresponds to 0, which is the minimum number of cases attended in the process, and the upper bound is theoretically limited by the capacity of the process. The parameter that most significantly affects data trend capture is changepoint-prior-scale, which determines how much the trend changes at the trend change points. This parameter needs tuning, since a too low value may cause under-fitting while a too high value may cause over-fitting. Accordingly, for this parameter, DeepSimulator explores values in the interval [0.001, 0.5]. Analogously, the parameter that most directly affects the seasonality capture is the seasonality-prior-scale. This parameter affects the flexibility of seasonality learning. If the value is too small, the model tends to focus on small fluctuations, while a large value may cause the model to focus only on large fluctuations. For this parameter, DeepSimulator explores values in [0.01, 10]. We do not define the Holidays component in the Prophet model. The holidays component could be discovered by a calendar discovery technique such as the one proposed in \cite{Estrada-Torres2021}, but discovering such calendars is orthogonal to the focus of the present paper.

We use grid search for selecting the best hyperparameters of the Prophet model, as the search space consists of only sixteen configurations (cf.\  Sect.~\ref{sec:exp_1}). We rely on the internal mechanisms embedded in Prophet for cross-validation and selection of cutoff points. During the simulation, we use the trained Prophet model to determine the number of cases to be created at each hour of the simulation. We then generate the start-times, for each simulation hour, by modeling the intercase arrival times via a normal distribution (within the hour).

\paragraph{Phase 3: activity timestamps generation.} \label{sec:hb_tool_phase3}
We enhance the activity sequences generated in Phase 1 to capture waiting times and processing times in this phase. The DeepSimulator trains two LSTM~\footnote{We used LSTM networks as the core of our predictive models since they are a well-known and proven technology to handle sequences, as the nature of a business process event log~\cite{Rama-Maneiro2020}. The proposed models were based on previous works~\cite{Tax2017, Camargo2019, Evermann2017} that have extensively explored several architectural options of LSTMs, such as the use of stacked vs. unstacked models or the use of shared layers vs. specialized ones.} models to perform two predictive tasks: the processing time of a given activity (herein called the \emph{current activity}) and the waiting time until the start of the next activity. This task differs in two ways from approaches used to predict the next event and its timestamp~\cite{Camargo2019, Taymouri2020}. First, we do not seek to predict the next event, since the sequences of activities are generated by the stochastic process model (cf.\ Phase 1). Second, we need to support changes in the process model (e.g., adding or removing tasks) for enabling what-if analysis.

\begin{wrapfigure}[8]{r}{0.61\textwidth}
    \vspace{-8mm}
    \includegraphics[width=0.6\textwidth]{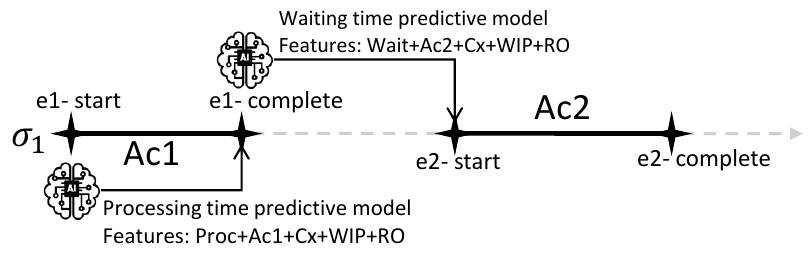}
    \caption{Predictive models timeline and features}
    \label{fig:pred_mod_timing}
\end{wrapfigure}

Therefore, we train one model specialized in predicting the processing time of the current activity and another specialized in predicting the waiting time until the next activity. Both models differ in the set of features since they act at different moments in the predictive phase, as shown in Fig.~\ref{fig:pred_mod_timing}. The processing time predictive model uses the following features as inputs: the label and processing time of the current activity, the time of day of the current activity's start timestamp, the day of the week, and inter-case features such as the Work-in-progress (WIP) of the process and the activity and Resources’ Occupation (RO) at the start of the activity. The waiting time predictive model uses the following features as inputs: the next activity's label, the time of day of the current activity's end timestamp, the day of the week, and inter-case features such as the WIP of the process and the RO at the end of the current activity.

In the \emph{log replay} step, we calculate the waiting and processing times of each activity by replaying each trace in the input log (or in a training subset thereof) against the process model discovered in Phase 1. An activity's processing time is the difference between its end and start timestamps. An activity's waiting time is the difference between its start time and enablement time, i.e., when it was ready to be executed according to the process model.  All waiting and processing times are scaled to the range [0...1] by dividing them by the largest values. 

In the \emph{feature engineering} step, we compute and encode all the remaining features used by the models. We calculate the time of the day as the elapsed seconds from the closest midnight until the event timestamp; this feature is scale over 86400 seconds. The day of the week is modeled as a categorical attribute and encoded using one-hot encoding. We include these latter features since they provide contextual information, allowing the model to find seasonal patterns in the data that may affect waiting and processing times. In the same way, considering that the overall process performance is affected by the process' WIP and the RO~\cite{laguna2018business}, we use two inter-case features that measure these variations. 

The WIP of the process measures the number of active tasks at each moment in the log transversally. The RO measures each resource pool's percentage occupancy in the log, implying that a new feature is created for each pool to record the occupation-specific variations. Since the information about the size and composition of the resource pools is not always included in the logs, we grouped resources into roles by using the algorithm described in~\cite{Song2008}. This algorithm discovers resource pools based on the definition of activity execution profiles for each resource and the creation of a correlation matrix of similarity of those profiles. WIP and RO are calculated by replaying over time the log events, recording the variations in both features at every time point. 

Finally, we encode the current activity's label using pre-trained embedded dimensions. We use embeddings for two reasons. First, embeddings help prevent exponential feature growth associated with one-hot encoding~\cite{Camargo2019}. Second, embedded dimensions allow adding new categories (i.e., \ activity labels) without altering the predictive model's structure. These embedded dimensions are an n-dimensional space, where each category (each activity level) is encoded as a point in that space. An independent network fed with positive and negative examples of associations between activities is used to map the activity labels to points. The network maps activities that co-occur or occur close to each other to nearby points. This mechanism also allows adding a new point in that space by updating the encoding model without altering the predictive model's input size. Each time a new activity is added to the process model for what-if analysis, we generate examples of traces involving this new activity and use these examples to determine the coordinates of the new activity label to be encoded in the embedded space. Then, we update the predictive model's embedded layers with the new definition, and the predictive model can handle the new activity label from that point on.
\begin{wrapfigure}[13]{r}{0.39\textwidth}
\vspace{-5mm}
    \includegraphics[width=0.35\textwidth]{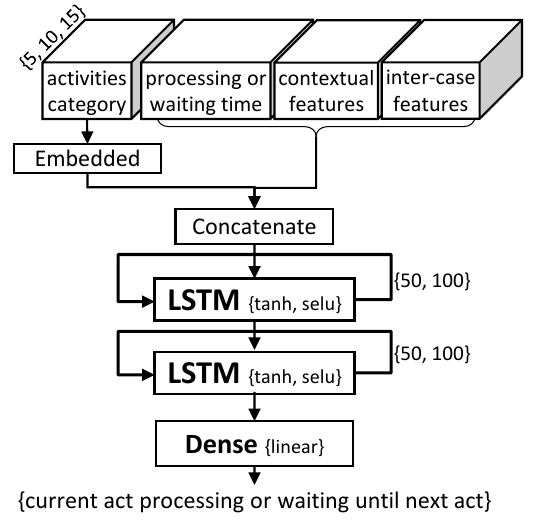}
    \caption{DL models architectures}
    \label{fig:models}
\end{wrapfigure}

Once encoded the features, we extract n-grams of fixed sizes from each trace to create the input sequences to train the model. As shown in Fig.~\ref{fig:models}, both models are composed of two stacked LSTM layers and a dense output. A model receives the sequences as inputs and the expected processing and waiting times as a target. The user can vary the number of units in the LSTM layers, the activation function, the size of the n-gram, and the use of all the RO inter-cases or just the one of the resource pool associated to the execution of the activity. 

\paragraph{Assembling the output log} The output log is generated by assembling each generated sequence (see Phase~1), with the generated case start time (see Phase~2) and the processing and waiting times predicted iteratively (see Phase~3). In each iteration, the trained model predicts times relative to the current activity in seconds, which are transformed into absolute times by adding them to the start time of the case. Then, the DeepSimulator generates a simulated log composed of a bag of traces, each trace consisting of a sequence of triplets (activity label, start-timestamp, end timestamp).
%
\section{Evaluation}
\label{sec:evaluation}

We empirically compare the DeepSimulator method vs. DDS and DL approaches in terms of the similarity of the simulated logs they generate relative to a fold of the original log. We also evaluate the accuracy of DeepSimulator for ``what-if" analysis tasks of modifying the case creation intesity and adding new activities to a process. 

\subsection{Datasets}

We evaluated the approaches using 9 logs that contain both start and end timestamps. We use real-life logs (R) from public and private sources and synthetic logs (S) generated from simulation models of real processes. Table~\ref{tab:eventlogs} provides descriptive statistics of the logs. The  BPI17W log have the largest number of traces and events, while CFS and P2P have fewer traces but more events/trace.

\begin{table}[ht]
\centering
\resizebox{\textwidth}{!}{%
\begin{tabular}{@{}cclrrrrrrl@{}}
\toprule
\textbf{Size} & \textbf{\begin{tabular}[c]{@{}c@{}}Source\end{tabular}} & \multicolumn{1}{c}{\textbf{Log}} & \multicolumn{1}{c}{\textbf{\#Traces}} & \multicolumn{1}{c}{\textbf{\#Events}} & \multicolumn{1}{c}{\textbf{\#Act.}} & \multicolumn{1}{c}{\textbf{\begin{tabular}[c]{@{}c@{}}Avg.\\ activities\\ per trace\end{tabular}}} & \multicolumn{1}{c}{\textbf{\begin{tabular}[c]{@{}c@{}}Avg.\\ duration\end{tabular}}} & \multicolumn{1}{c}{\textbf{\begin{tabular}[c]{@{}c@{}}Max.\\ duration\end{tabular}}} & \multicolumn{1}{c}{\textbf{Description}} \\ \midrule
\multicolumn{1}{c|}{\multirow{4}{*}{\rotatebox[origin=c]{90}{LARGE}}} & \multicolumn{1}{c|}{R} & \multicolumn{1}{l|}{BPI17W} & \multicolumn{1}{r|}{30276} & \multicolumn{1}{r|}{240854} & \multicolumn{1}{r|}{8} & \multicolumn{1}{r|}{7.96} & \multicolumn{1}{r|}{12.66 days} & \multicolumn{1}{r|}{286.07 days} & \begin{tabular}[c]{@{}l@{}}Dutch financial institution updated\end{tabular} \\ \cmidrule(l){2-10} 
\multicolumn{1}{c|}{} & \multicolumn{1}{c|}{R} & \multicolumn{1}{l|}{BPI12W} & \multicolumn{1}{r|}{8616} & \multicolumn{1}{r|}{59302} & \multicolumn{1}{r|}{6} & \multicolumn{1}{r|}{6.88} & \multicolumn{1}{r|}{8.91 days} & \multicolumn{1}{r|}{85.87 days} & \begin{tabular}[c]{@{}l@{}}Dutch financial institution\end{tabular} \\ \cmidrule(l){2-10} 
\multicolumn{1}{c|}{} & \multicolumn{1}{c|}{S} & \multicolumn{1}{l|}{CVS} & \multicolumn{1}{r|}{10000} & \multicolumn{1}{r|}{103906} & \multicolumn{1}{r|}{15} & \multicolumn{1}{r|}{10.39} & \multicolumn{1}{r|}{7.58 days} & \multicolumn{1}{r|}{21.0 days} & \begin{tabular}[c]{@{}l@{}}CVS retail pharmacy**\end{tabular} \\ \cmidrule(l){2-10} 
\multicolumn{1}{c|}{} & \multicolumn{1}{c|}{S} & \multicolumn{1}{l|}{CFM} & \multicolumn{1}{r|}{1670} & \multicolumn{1}{r|}{44373} & \multicolumn{1}{r|}{29} & \multicolumn{1}{r|}{26.57} & \multicolumn{1}{r|}{0.76 days} & \multicolumn{1}{r|}{5.83 days} & \begin{tabular}[c]{@{}l@{}}Anonymized onfidential process**\end{tabular} \\ \midrule
\multicolumn{1}{c|}{\multirow{5}{*}{\rotatebox[origin=c]{90}{SMALL}}} & \multicolumn{1}{c|}{R} & \multicolumn{1}{l|}{INS} & \multicolumn{1}{r|}{1182} & \multicolumn{1}{r|}{23141} & \multicolumn{1}{r|}{9} & \multicolumn{1}{r|}{19.58} & \multicolumn{1}{r|}{70.93 days} & \multicolumn{1}{r|}{599.9 days} & \begin{tabular}[c]{@{}l@{}}Insurance claims process*\end{tabular} \\ \cmidrule(l){2-10} 
\multicolumn{1}{c|}{} & \multicolumn{1}{c|}{R} & \multicolumn{1}{l|}{ACR} & \multicolumn{1}{r|}{954} & \multicolumn{1}{r|}{4962} & \multicolumn{1}{r|}{16} & \multicolumn{1}{r|}{5.2} & \multicolumn{1}{r|}{14.89 days} & \multicolumn{1}{r|}{135.84 days} & \begin{tabular}[c]{@{}l@{}}Academic Credential Recognition\end{tabular} \\ \cmidrule(l){2-10} 
\multicolumn{1}{c|}{} & \multicolumn{1}{c|}{R} & \multicolumn{1}{l|}{MP} & \multicolumn{1}{r|}{225} & \multicolumn{1}{r|}{4503} & \multicolumn{1}{r|}{24} & \multicolumn{1}{r|}{20.01} & \multicolumn{1}{r|}{20.63 days} & \multicolumn{1}{r|}{87.5 days} & \begin{tabular}[c]{@{}l@{}}Manufacturing Production\end{tabular} \\ \cmidrule(l){2-10} 
\multicolumn{1}{c|}{} & \multicolumn{1}{c|}{S} & \multicolumn{1}{l|}{CFS} & \multicolumn{1}{r|}{800} & \multicolumn{1}{r|}{21221} & \multicolumn{1}{r|}{29} & \multicolumn{1}{r|}{26.53} & \multicolumn{1}{r|}{0.83 days} & \multicolumn{1}{r|}{4.09 days} & \begin{tabular}[c]{@{}l@{}}Anonymized confidential process**\end{tabular} \\ \cmidrule(l){2-10} 
\multicolumn{1}{c|}{} & \multicolumn{1}{c|}{S} & \multicolumn{1}{l|}{P2P} & \multicolumn{1}{r|}{608} & \multicolumn{1}{r|}{9119} & \multicolumn{1}{r|}{21} & \multicolumn{1}{r|}{15} & \multicolumn{1}{r|}{21.46 days} & \multicolumn{1}{r|}{108.31 days} & \begin{tabular}[c]{@{}l@{}}Purchase-to-Pay process\end{tabular} \\ \bottomrule
\end{tabular}%
}
\caption{Event logs description. (*) Private logs, (**) Generated  from  simulation models of real processes}
\vspace{-10mm}
\label{tab:eventlogs}
\end{table}

\subsection{Evaluation measures}
\label{sec:evalmeasures}

To evaluate the accuracy of a model $M$ produced by one of the methods under evaluation, we compute a distance measure between a log generated by model $M$ and a ground-truth log (a testing subset of the original log). In all of our experiments, we use two distance measures: the Mean Absolute Error (MAE) of cycle times and the Earth-Mover's Distance (EMD) of the normalized histograms of activity timestamps grouped by day/hour.

The \emph{cycle time MAE} measures the temporal similarity between two logs at the \textit{trace level}. The absolute error of a pair of traces T1 and T2 is the absolute value of the difference between their cycle times. The cycle time MAE is the mean of the absolute errors over a collection of paired traces. Given this trace distance notion, we pair each trace in the generated log with a trace in the original log using the Hungarian algorithm~\cite{Kuhn1955} so that the sum of the trace errors between the paired traces is minimal.

The cycle time MAE is a rough measure of the temporal similarity between the ground-truth and the simulated traces. But it does not consider the start time of each case, nor the start and end timestamps of each activity. To complement MAE, we use the \emph{Earth Mover's Distance (EMD)} between the normalized histograms of the timestamps grouped by day/hour in the ground-truth and the generated logs. The EMD between two histograms, H1 and H2, is the minimum number of units that need to be added, removed, or transferred across columns in H1 to transform it into H2. The EMD is zero if the observed distributions in the two logs are identical, and it tends to one the more they differ.

\subsection{Experiment 1: AS-IS Accuracy of generated models} 
\label{sec:exp_1}

\subsubsection{Setup}
\begin{figure}[ht]
  \centering
    \includegraphics[width=\textwidth]{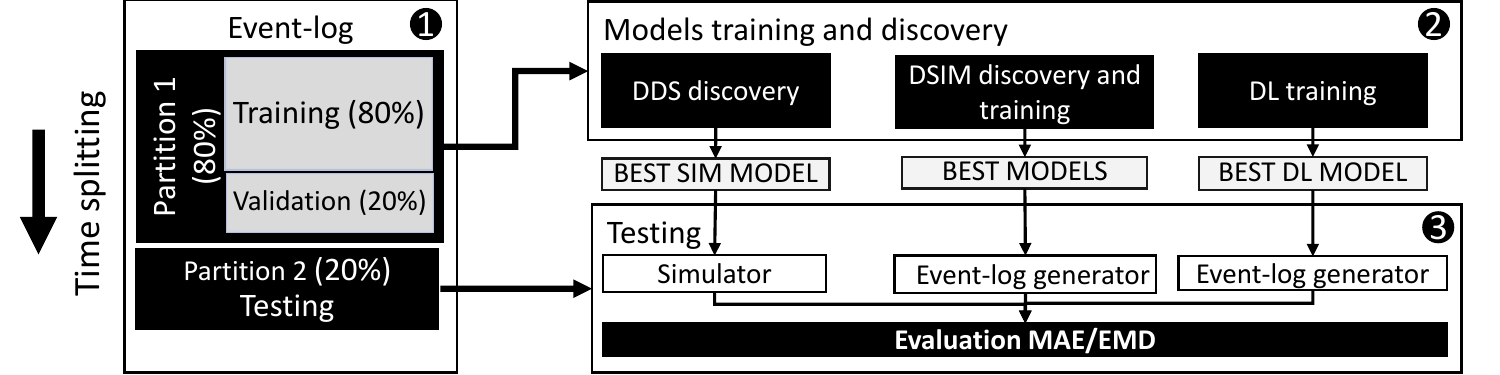}
    \caption{Setup of experiment 1}
    \label{fig:exp_pipeline}
    \vspace*{-5mm}
\end{figure}

This experiment aims to compare the accuracy of DeepSimulator models (herein called DSIM models) vs. DDS and DL models. We use SIMOD~\cite{Camargo2020} as a baseline DDS approach since it is fully automated both w.r.t.\ parameter discovery and tuning. As DL baselines, we use an adaptation of the LSTM approach proposed by Camargo~et~al.~\cite{Camargo2019} (herein labeled the LSTM method) as well as the GAN-LSTM approach by Taymouri~et~al.~\cite{Taymouri2020} (herein labeled GAN). Both of these DL approaches have been shown to achieve high accuracy w.r.t.\ the task of generating timestamped trace suffixes~\cite{Rama-Maneiro2020}. 
Fig.~\ref{fig:exp_pipeline} summarizes the experimental setup.
We use the hold-out method with a temporal split criterion to divide the logs into two main folds: 80\% for training-validation and 20\% for testing. From the first fold, we took the first 80\% for training and 20\% for validation. We use temporal splits to prevent information leakage \cite{Camargo2019,Taymouri2020}. 

The DDS technique (SIMOD) is set to explore 15 parameter configurations
to tune the stochastic process model. For each configuration, we execute five simulation runs and compute the CFLS measure (cf.~Section~\ref{sec:hb_tool_phase1}) between each simulated log and the validation fold. We select the stochastic model that gives the lowest average CFLS w.r.t.\ the validation fold. The optimizer is set to explore 20 simulation parameter configurations (i.e.\ the parameters that Simod uses to model resources and processing times), again using five simulation runs per configuration. We select the configuration with the lowest average EMD (cf.\ Section~\ref{sec:evalmeasures}) between the simulated log and the validation fold. We used the parameter ranges given in Table~\ref{tab:seach_space} for tuning.

The LSTM technique is hyperparameter-optimized using grid search over a space of 48 possible configurations (see Table~\ref{tab:seach_space})
For LSTM model training, we use 200 epochs, the cycle time MAE as the model's loss function, Nadam as the optimizer, and early stopping and dropout to avoid model over-training. 
The GAN technique is configured to dynamically adjust the size of the hidden units in each layer so that their size is twice the input's size, as proposed by the authors~\cite{Taymouri2020}. We use 25 training epochs, a batch of size five, and a prefix size of five. 
DSIM is tuned by randomly exploring 15 parameter configurations with five simulation runs per configuration in the stochastic model discovery phase (cf.~Section~\ref{sec:tool}, Phase 1). In Phases 2 and 3, we use grid search to explore the space of  hyperparameter configurations specified in Table~\ref{tab:seach_space}

We generate four models per log: one SIMOD, one LSTM, one GAN, and one DSIM.  
We then generate five logs per retained model, each with the same number of traces as the original log's testing fold to ensure the comparability. Each generated log is compared with the testing fold using the MAE and EMD measures. We report the mean of each of these measures across 5 runs.

\begin{table}[ht]
\centering
\resizebox{\textwidth}{!}{%
\begin{tabular}{@{}lllll@{}}
\toprule
\textbf{Model} & \textbf{Stage} & \textbf{Parameter} & \textbf{Distribution} & \textbf{Values} \\ \midrule
\multicolumn{1}{l|}{\multirow{9}{*}{\rotatebox[origin=c]{90}{SIMOD}}} & \multicolumn{1}{l|}{\multirow{3}{*}{\begin{tabular}[c]{@{}l@{}}Structure \\ discovery\end{tabular}}} & \multicolumn{1}{l|}{Parallelism threshold   ($\epsilon$)} & \multicolumn{1}{l|}{Uniform} & {[}0...1{]} \\ \cmidrule(l){3-5} 
\multicolumn{1}{l|}{} & \multicolumn{1}{l|}{} & \multicolumn{1}{l|}{Percentile for frequency threshold ($\eta$)} & \multicolumn{1}{l|}{Uniform} & {[}0...1{]} \\ \cmidrule(l){3-5} 
\multicolumn{1}{l|}{} & \multicolumn{1}{l|}{} & \multicolumn{1}{l|}{Conditional branching probabilities} & \multicolumn{1}{l|}{Categorical} & \{Equiprobable,   Discovered\} \\ \cmidrule(l){2-5} 
\multicolumn{1}{l|}{} & \multicolumn{1}{l|}{\multirow{6}{*}{\begin{tabular}[c]{@{}l@{}}Time-related \\ parameters \\ discovery\end{tabular}}} & \multicolumn{1}{l|}{Log repair technique} & \multicolumn{1}{l|}{Categorical} & \{Repair, Removal,   Replace\} \\ \cmidrule(l){3-5} 
\multicolumn{1}{l|}{} & \multicolumn{1}{l|}{} & \multicolumn{1}{l|}{Resource pools similarity threshold} & \multicolumn{1}{l|}{Uniform} & {[}0...1{]} \\ \cmidrule(l){3-5} 
\multicolumn{1}{l|}{} & \multicolumn{1}{l|}{} & \multicolumn{1}{l|}{Resource availability calendar support} & \multicolumn{1}{l|}{Uniform} & {[}0...1{]} \\ \cmidrule(l){3-5} 
\multicolumn{1}{l|}{} & \multicolumn{1}{l|}{} & \multicolumn{1}{l|}{Resource availability calendar confidence} & \multicolumn{1}{l|}{Uniform} & {[}0...1{]} \\ \cmidrule(l){3-5} 
\multicolumn{1}{l|}{} & \multicolumn{1}{l|}{} & \multicolumn{1}{l|}{Instances creation calendar support} & \multicolumn{1}{l|}{Uniform} & {[}0...1{]} \\ \cmidrule(l){3-5} 
\multicolumn{1}{l|}{} & \multicolumn{1}{l|}{} & \multicolumn{1}{l|}{Instances creation calendars confidence} & \multicolumn{1}{l|}{Uniform} & {[}0...1{]} \\ \midrule
\multicolumn{1}{l|}{\multirow{5}{*}{\rotatebox[origin=c]{90}{LSTM}}} & \multicolumn{1}{l|}{\multirow{5}{*}{Training}} & \multicolumn{1}{l|}{N-gram size} & \multicolumn{1}{l|}{Categorical} & {[}5, 10, 15{]} \\ \cmidrule(l){3-5} 
\multicolumn{1}{l|}{} & \multicolumn{1}{l|}{} & \multicolumn{1}{l|}{Input scaling method} & \multicolumn{1}{l|}{Categorical} & \{Max, Lognormal\} \\ \cmidrule(l){3-5} 
\multicolumn{1}{l|}{} & \multicolumn{1}{l|}{} & \multicolumn{1}{l|}{\# units in hidden layer} & \multicolumn{1}{l|}{Categorical} & \{50, 100\} \\ \cmidrule(l){3-5} 
\multicolumn{1}{l|}{} & \multicolumn{1}{l|}{} & \multicolumn{1}{l|}{Activation function for hidden layers} & \multicolumn{1}{l|}{Categorical} & \{selu, tanh\} \\ \cmidrule(l){3-5} 
\multicolumn{1}{l|}{} & \multicolumn{1}{l|}{} & \multicolumn{1}{l|}{Model type} & \multicolumn{1}{l|}{Categorical} & \{shared\_cat,   concatenated\} \\ \midrule
\multicolumn{1}{l|}{\multirow{9}{*}{\rotatebox[origin=c]{90}{DSIM}}} & \multicolumn{1}{l|}{\multirow{3}{*}{\begin{tabular}[c]{@{}l@{}}Structure \\ generation\end{tabular}}} & \multicolumn{1}{l|}{Parallelism threshold   ($\epsilon$)} & \multicolumn{1}{l|}{Uniform} & {[}0...1{]} \\ \cmidrule(l){3-5} 
\multicolumn{1}{l|}{} & \multicolumn{1}{l|}{} & \multicolumn{1}{l|}{Percentile for frequency threshold ($\eta$)} & \multicolumn{1}{l|}{Uniform} & {[}0...1{]} \\ \cmidrule(l){3-5} 
\multicolumn{1}{l|}{} & \multicolumn{1}{l|}{} & \multicolumn{1}{l|}{Conditional branching probabilities} & \multicolumn{1}{l|}{Categorical} & \{Equiprobable,   Discovered\} \\ \cmidrule(l){2-5} 
\multicolumn{1}{l|}{} & \multicolumn{1}{l|}{\multirow{2}{*}{\begin{tabular}[c]{@{}l@{}}Cases start-times \\ generation\end{tabular}}} & \multicolumn{1}{l|}{changepoint-prior-scale} & \multicolumn{1}{l|}{Categorical} & \{0.001, 0.01, 0.1,   0.5\} \\ \cmidrule(l){3-5} 
\multicolumn{1}{l|}{} & \multicolumn{1}{l|}{} & \multicolumn{1}{l|}{seasonality-prior-scale} & \multicolumn{1}{l|}{Categorical} & \{0.01, 0.1, 1.0,   10.0\} \\ \cmidrule(l){2-5} 
\multicolumn{1}{l|}{} & \multicolumn{1}{l|}{\multirow{4}{*}{\begin{tabular}[c]{@{}l@{}}Timestamps \\ generation\end{tabular}}} & \multicolumn{1}{l|}{N-gram size} & \multicolumn{1}{l|}{Categorical} & \{5, 10, 15\} \\ \cmidrule(l){3-5} 
\multicolumn{1}{l|}{} & \multicolumn{1}{l|}{} & \multicolumn{1}{l|}{\# units in hidden layer} & \multicolumn{1}{l|}{Categorical} & \{50, 100\} \\ \cmidrule(l){3-5} 
\multicolumn{1}{l|}{} & \multicolumn{1}{l|}{} & \multicolumn{1}{l|}{Activation function for hidden layers} & \multicolumn{1}{l|}{Categorical} & \{selu, tanh\} \\ \cmidrule(l){3-5} 
\multicolumn{1}{l|}{} & \multicolumn{1}{l|}{} & \multicolumn{1}{l|}{Single resource-pool intercase feature} & \multicolumn{1}{l|}{Boolean} & \{True, False\} \\ \bottomrule
\end{tabular}%
}
\caption{Hyperparameters used by optimization techniques}
\label{tab:seach_space}
\end{table}

\vspace*{-2mm}

\subsubsection{Results}

\begin{wraptable}[13]{r}{0.68\textwidth}
\centering
\resizebox{0.65\textwidth}{!}{%
\begin{tabular}{@{}lll|r|r|r|r|r|r@{}}
\toprule
\multicolumn{1}{c}{\multirow{2}{*}{\textbf{Size}}} &
  \multicolumn{1}{c}{\multirow{2}{*}{\textbf{Type}}} &
  \multicolumn{1}{c|}{\multirow{2}{*}{\textbf{Log}}} &
  \multicolumn{4}{c|}{\textbf{MAE}} &
  \multicolumn{2}{c}{\textbf{EMD}} \\ \cmidrule(l){4-9} 
\multicolumn{1}{c}{} &
  \multicolumn{1}{c}{} &
  \multicolumn{1}{c|}{} &
  \multicolumn{1}{c|}{\textbf{GAN}} &
  \multicolumn{1}{c|}{\textbf{LSTM}} &
  \multicolumn{1}{c|}{\textbf{SIMOD}} &
  \multicolumn{1}{c|}{\textbf{DSIM}} &
  \multicolumn{1}{c|}{\textbf{SIMOD}} &
  \multicolumn{1}{c}{\textbf{DSIM}} \\ \midrule
\multirow{4}{*}{\rotatebox[origin=c]{90}{LARGE}} &
  \multirow{2}{*}{R} &
  BPI17W &
  828165 &
  603688 &
  961727 &
  {\ul \textit{418422}} &
  {\ul \textit{0.016873}} &
  0.034584 \\ \cmidrule(l){3-9} 
 &
   &
  BPI12W &
  653656 &
  {\ul \textit{327350}} &
  662333 &
  548813 &
  0.056789 &
  {\ul \textit{0.020098}} \\ \cmidrule(l){2-9} 
 &
  \multirow{2}{*}{S} &
  CVS &
  952004 &
  667715 &
  1067258 &
  {\ul \textit{158902}} &
  0.018955 &
  {\ul \textit{0.000004}} \\ \cmidrule(l){3-9} 
 &
   &
  CFM &
  956289 &
  15078 &
  252458 &
  {\ul \textit{8441}} &
  0.240567 &
  {\ul \textit{0.239087}} \\ \midrule
\multirow{5}{*}{\rotatebox[origin=c]{90}{SMALL}} &
  \multirow{3}{*}{R} &
  INS &
  1302337 &
  1516368 &
  {\ul \textit{1090179}} &
  1190019 &
  0.143675 &
  {\ul \textit{0.142097}} \\ \cmidrule(l){3-9} 
 &
   &
  ACR &
  296094 &
  341694 &
  230363 &
  {\ul \textit{165411}} &
  0.207050 &
  {\ul \textit{0.205106}} \\ \cmidrule(l){3-9} 
 &
   &
  MP &
  210714 &
  321147 &
  298641 &
  {\ul \textit{157453}} &
  0.062227 &
  {\ul \textit{0.050479}} \\ \cmidrule(l){2-9} 
 &
  \multirow{2}{*}{S} &
  CFS &
  717266 &
  33016 &
  {\ul \textit{15297}} &
  24326 &
  {\ul \textit{0.222515}} &
  0.266749 \\ \cmidrule(l){3-9} 
 &
   &
  P2P &
  2347070 &
  2495593 &
  1892415 &
  {\ul \textit{1836863}} &
  {\ul \textit{0.130655}} &
  0.132898 \\ \bottomrule
\end{tabular}%
}
\vspace{-2mm}
\caption{Evaluation results (lower values are better)}
\label{tab:results}
\end{wraptable}
Table~\ref{tab:results} show the results grouped by metrics, log size and source type. Note that MAE and EMD are error/distance measures (lower is better).
In 3 out of 4 large logs, DSIM outperforms LSTM and SIMOD w.r.t the MAE measure. In the small logs, DSIM attains lower MAE in 3 out of 5 logs and has similar MAE w.r.t.\ SIMOD in one other log. 
Similarly, DSIM outperforms SIMOD in 6 of 9 logs w.r.t.\ EMD, and achieves results similar to SIMOD in two others.\footnote{We cannot measure EMD for the LSTM and GAN models because EMD requires that the timestamps in the log are absolute timestamps, while the LSTM and GAN approaches produce relative timestamps (w.r.t.\ an unknown case start-time). It would be possible to extend the LSTM/GAN approaches to generate logs with absolute timestamps by coupling them with a model to generate start-times (e.g.\ based on Prophet) but this extension is outside the scope of this paper.}
%

The results suggest that DSIM is often able to outperform the baselines when it comes to replicating the as-is behavior recorded in an event log. This conclusion should be tempered in light of two threats to validity: (i) an external threat to validity stemming from the limited number of events logs in the experiment; and (ii) a threat to construct validity created by the fact that the accuracy measures do not necessarily capture all the nuances in the control-flow and temporal behavior captured in the original and simulated event logs.

\subsection{Experiment 2: What-if analysis}

In this experiment, we compare DSIM's ability to simulate a process after a change (what-if analysis). We consider two scenarios. In the first one, we assess DSIM's ability to capture variations in the inter-arrival time between cases (a.k.a.\emph{arrival intensity}), specifically alternations between periods of lower arrival intensity and periods of higher intensity. In the second experiment, we evaluate the ability of DSIM (vs baselines) to estimate the impact of adding a never-before-observed activity to a process. This scenario is challenging for the DSIM and the DL models because these models need to infer the temporal behavior of the new activity using their embedding layers.


\vspace*{-2mm}
\subsubsection{Setup scenario 1} We create two modified versions of the three largest records (BPI12W, BPI17W, and CVS). First, to capture a periodic reduction in the arrival intensity of cases, we divide each log into six batches of the same number of cases, and then we create two alternating groups of three batches each.
\begin{wrapfigure}[10]{R}{0.55\textwidth}
    \includegraphics[width=0.53\textwidth]{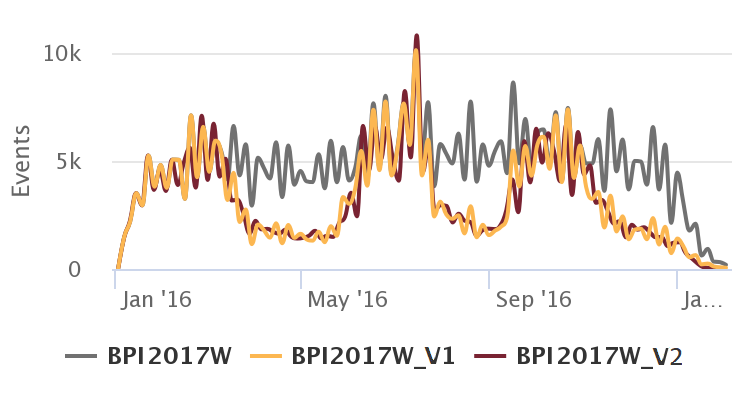}
    \vspace*{-3mm}
    \caption{Original vs modified case creations}
\label{fig:hb_modif_log}
\end{wrapfigure}

The first group consists of batches 1, 3, and 5. The batches in this group are left unaltered. These groups represent periods of high arrival intensity. The second group comprises batches 2, 4, and 6. This group is used to emulate periods of low arrival intensity. To do so, we reduce the case arrival rate in these batches by 1/3, by randomly eliminating two out of three cases. 
The above altered version of the log capture a situation where the arrival intensity varies, but the waiting times within the cases remain the same. In general, when the arrival intensity goes down, the waiting times should go down. To capture this latter scenario, we create a second altered version of the log to capture decreases in waiting times associated with decreases in arrival rates. Accordingly, we take the first altered log, and we reduce the waiting times in group 2 (batches 2, 4, and 6) by 30\%. For illustration, Fig.~\ref{fig:hb_modif_log} sketches the modifications made to the BPI17W log. 

After altering the logs as above, we train and evaluate the DSIM and SIMOD models as in Experiment~1 (see Fig.~\ref{fig:exp_pipeline}). Since we are particularly interested here in assessing the ability of the evaluated techniques to model the temporal dynamics of case arrivals, we also report the Dynamic Time Warping (DTW) distance between the time series of the number of cases generated per hour of the day, between the testing partition and the logs generated. 


\begin{wrapfigure}[10]{rt}{0.53\textwidth}
\vspace*{-7mm}
    \includegraphics[width=0.52\textwidth]{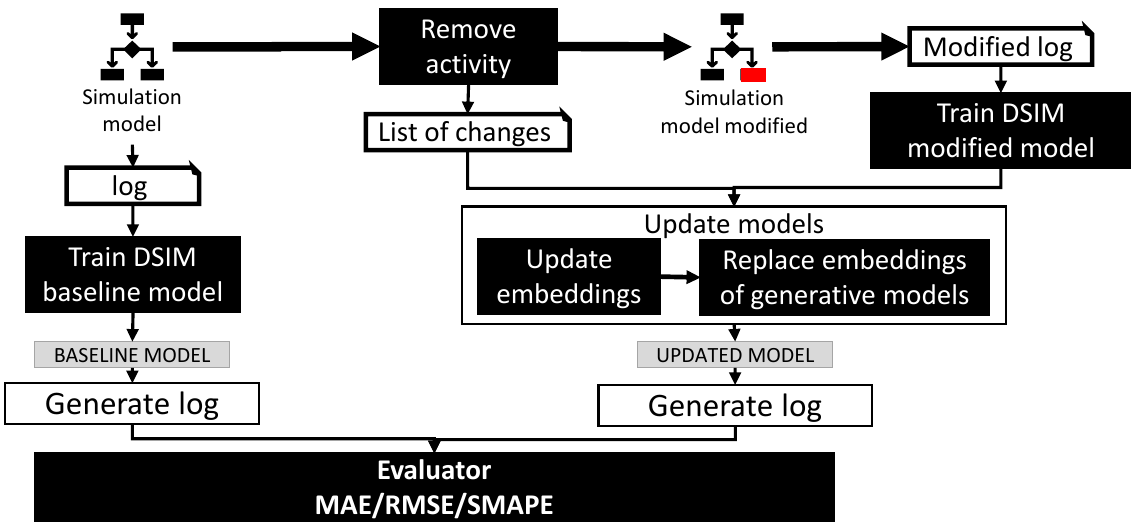}
    \caption{Pipeline of scenario 2}
    \label{fig:setup_2}
\end{wrapfigure}

\vspace*{-2mm}
\subsubsection{Setup scenario 2} For each of the synthetic logs (CVS and CFM), we select a random activity A and eliminate all its occurrences from the log. We then train a DSIM simulation model using this modified log (cf.\ left-hand side of Fig.~\ref{fig:setup_2}b). Next, we generate synthetic data consisting of positive and negative samples of pairs composed by the activity label and associated resource (cf.\ Sec.~\ref{sec:tool}). Using these synthetic samples, we update the embedded dimensions to include activity A (without modifying the embedding of the remaining activities). We then plug the updated embedding into the previously trained DSIM model (cf.\ right-hand side of Fig.~\ref{fig:setup_2}). We calculate the errors of the DSIM model of the ``as-is'' process (before a change) and the DSIM model of the ``what-if'' process (after adding an activity). 
We measure the error using MAE. Additionally, we report the RMSE and SMAPE metrics to confirm that the results do not depend on the chosen metric.

\vspace*{-2mm}
\subsubsection{Results scenario 1}
Table~\ref{tab:exp2_results} presents the MAE, EMD and DTW results. DSIM has a lower error in cycle times in all cases. In version 2 the MAE logs are considerably reduced compared to those of the version 1. We can explain this result because the DL models in charge of predicting waiting and processing times consider inter-case attributes that capture the workload of the process, allowing their adjustment to workload variations. Regarding the results of EMD and DTW, both metrics follow the same trends. In most cases, DSIM obtains the best results. This trend is more evident in version 2, in which DSIM gets better results in all cases. These results indicate that Prophet and DL models are more effective than the baselines at capturing the temporal variations in waiting times due to a decrease in the arrival intensity. This observation should again be tempered by the threats to validity acknowledged above.


\enlargethispage{.5\baselineskip}

\vspace*{-2mm}
\subsubsection{Results scenario 2}
Table~\ref{tab:exp2_results} presents the MAE, RSME, and SMAPE grouped for each log, both for the simulation model of the as-is process vs. the model derived after the addition of the activity (what-if model). The what-if model has higher MAE than the baseline models in both event logs. The higher error values are evident in the CVS log, where the SMAPE of the updated model is 184\% compared to 31.97\% for the baseline model. These results suggest that embedded dimensions incorporated in DSIM can predict the presence of activities that were not present in the training set, but it is unable to adequately estimate their temporal behavior. This observation suggests that DSIM could be extended with more sophisticated embedding techniques (e.g.\ word2vec or transformer models) to better capture the temporal dynamics of previously unobserved activities (by analogy to activities that have been observed in similar contexts).


\begin{table}[ht]
\centering
\resizebox{0.75\textwidth}{!}{%
\begin{tabular}{@{}l|lrrrrrr@{}}
\toprule
\multirow{10}{*}{\rotatebox[origin=c]{90}{\textbf{Scenario 1}}} &
  \multicolumn{1}{c|}{\multirow{2}{*}{\textbf{Log}}} &
  \multicolumn{2}{c|}{\textbf{MAE}} &
  \multicolumn{2}{c|}{\textbf{EMD}} &
  \multicolumn{2}{c}{\textbf{DTW}} \\  
 &
  \multicolumn{1}{c|}{} &
  \multicolumn{1}{c}{\textbf{SIMOD}} &
  \multicolumn{1}{c|}{\textbf{DSIM}} &
  \multicolumn{1}{c}{\textbf{SIMOD}} &
  \multicolumn{1}{c|}{\textbf{DSIM}} &
  \multicolumn{1}{c}{\textbf{SIMOD}} &
  \multicolumn{1}{c}{\textbf{DSIM}} \\ \cmidrule(l){2-8} 
 &
  \multicolumn{7}{l}{\textit{Version 1}} \\
 &
  \multicolumn{1}{l|}{BPI17W} &
  971151 &
  \multicolumn{1}{r|}{{\ul \textit{\textbf{417572}}}} &
  {\ul \textit{\textbf{0.02222}}} &
  \multicolumn{1}{r|}{0.03593} &
  {\ul \textit{\textbf{3185}}} &
  3647 \\
 &
  \multicolumn{1}{l|}{BPI12W} &
  660211 &
  \multicolumn{1}{r|}{{\ul \textit{\textbf{534341}}}} &
  0.11295 &
  \multicolumn{1}{r|}{{\ul \textit{\textbf{0.04853}}}} &
  515 &
  {\ul \textit{\textbf{458}}} \\
 &
  \multicolumn{1}{l|}{CVS} &
  1489252 &
  \multicolumn{1}{r|}{{\ul \textit{\textbf{467572}}}} &
  0.03213 &
  \multicolumn{1}{r|}{{\ul \textit{\textbf{0.00001}}}} &
  3380 &
  {\ul \textit{\textbf{849}}} \\ \cmidrule(l){2-8} 
 &
  \multicolumn{7}{l}{\textit{Version 2}} \\
 &
  \multicolumn{1}{l|}{BPI17W} &
  895524 &
  \multicolumn{1}{r|}{{\ul \textit{\textbf{290980}}}} &
  0.06438 &
  \multicolumn{1}{r|}{{\ul \textit{\textbf{0.03218}}}} &
  4528 &
  {\ul \textit{\textbf{3431}}} \\
 &
  \multicolumn{1}{l|}{BPI12W} &
  550266 &
  \multicolumn{1}{r|}{{\ul \textit{\textbf{524995}}}} &
  0.25888 &
  \multicolumn{1}{r|}{{\ul \textit{\textbf{0.22003}}}} &
  726 &
  {\ul \textit{\textbf{507}}} \\
 &
  \multicolumn{1}{l|}{CVS} &
  540112 &
  \multicolumn{1}{r|}{{\ul \textit{\textbf{246159}}}} &
  0.15674 &
  \multicolumn{1}{r|}{{\ul \textit{\textbf{0.05708}}}} &
  2453 &
  {\ul \textit{\textbf{1967}}} \\ \midrule
\multirow{4}{*}{\rotatebox[origin=c]{90}{\textbf{Scenario 2}}} &
  \multicolumn{1}{c|}{\multirow{2}{*}{\textbf{Log}}} &
  \multicolumn{2}{c|}{\textbf{MAE}} &
  \multicolumn{2}{c|}{\textbf{RMSE}} &
  \multicolumn{2}{c}{\textbf{SMAPE}} \\
 &
  \multicolumn{1}{c|}{} &
  \multicolumn{1}{c}{\textbf{AS-IS}} &
  \multicolumn{1}{c|}{\textbf{WHAT-IF}} &
  \multicolumn{1}{c}{\textbf{AS-IS}} &
  \multicolumn{1}{c|}{\textbf{WHAT-IF}} &
  \multicolumn{1}{c}{\textbf{AS-IS}} &
  \multicolumn{1}{c}{\textbf{WHAT-IF}} \\ \cmidrule(l){2-8} 
 &
  \multicolumn{1}{l|}{CFM} &
  {\ul \textit{\textbf{7155}}} &
  \multicolumn{1}{r|}{17546} &
  {\ul \textit{\textbf{22006}}} &
  \multicolumn{1}{r|}{33137} &
  {\ul \textit{\textbf{0.15629}}} &
  0.28762 \\
 &
  \multicolumn{1}{l|}{CVS} &
  {\ul \textit{\textbf{283061}}} &
  \multicolumn{1}{r|}{1040344} &
  {\ul \textit{\textbf{357717}}} &
  \multicolumn{1}{r|}{1052255} &
  {\ul \textit{\textbf{0.31972}}} &
  1.84601 \\ \bottomrule
\end{tabular}%
}
\caption{Results of scenarios 1 and 2}
\vspace*{-15mm}
\label{tab:exp2_results}
\end{table}
%
\section{Conclusion}
\label{sec:conclusion}

This paper presented a method, namely DeepSimulator, to learn BPS models from event logs based on process mining and DL techniques. The design of DeepSimulator draws upon the observation that DDS methods (based on process mining) do not capture delays between activities caused by factors other than resource contention (e.g.\ fatigue, batching, inter-process dependencies). In contrast, DL techniques can learn temporal patterns without assuming these patterns stem only from resource contention. Accordingly, DeepSimulator discovers a stochastic process model from a log using process mining, and then uses a DL  model to add timestamps to the events produced by the stochastic model. The stochastic model can be modified (activities may be added/removed, branching probabilities may be altered), thus enabling some forms of what-if analysis.

\enlargethispage{\baselineskip}
The paper reported on an empirical comparison of the proposed technique with respect to: (i) its ability to replicate the observed as-is behavior; and (ii) its ability to estimate the impact of changes (what-if settings).
The evaluation in the ``as-is'' setting shows that the DeepSimulator method outperforms the baselines (one DDS and two DL methods). The evaluation in the what-if analysis setting shows that DeepSimulator can better estimate the impact of  changes in the arrival rate of new cases (the demand) in settings where such changes have been previously observed in the data. However, the accuracy of DeepSimulator degraded when evaluated in a previously unobserved scenario, specifically a scenario where a completely new task is added to the process. We foresee that this drawback could be at least partially addressed by adapting more sophisticated embedding techniques, such as word2vec or transformer models.

Another avenue for future work is to extend the approach to generate events with a ``resource'' attribute. A related avenue is to extend the approach to support a broader range of changes, such as adding or removing resources. Yet another avenue is to validate the proposed method via case studies, to complement the post-mortem evaluation reported in this paper.

\medskip\noindent\textbf{Reproducibility}
\label{sec:rep_pack}
The source code is available at \url{https://github.com/AdaptiveBProcess/DeepSimulator.git}. The datasets, models, and evaluation results can be found at \url{https://doi.org/10.5281/zenodo.5734443}.
\vspace*{-5mm}
%

\bibliography{bib/references.bib}
\bibliographystyle{IEEEtran}
\end{document}